\documentclass[sigconf]{acmart}
\usepackage{subcaption} 
\usepackage{multirow}
\AtBeginDocument{%
	}

\copyrightyear{2025}
\acmYear{2025}
\setcopyright{acmlicensed}
\acmConference[MM '25] {Proceedings of the 33rd ACM International Conference on Multimedia}{October 27--31, 2025}{Dublin, Ireland.}
\acmBooktitle{Proceedings of the 33rd ACM International Conference on Multimedia (MM '25), October 27--31, 2025, Dublin, Ireland}
\acmISBN{979-8-4007-2035-2/2025/10}
\acmDOI{10.1145/3746027.3754960}
\begin{document}

\title{PPJudge: Towards Human-Aligned Assessment of Artistic Painting Process}

%
\author{Shiqi Jiang}
\affiliation{%
	\department{School of Computer Science and Technology}
	\institution{East China Normal University}
	\city{Shanghai}
	\country{China}}
\email{52265901032@stu.ecnu.edu.cn}
\orcid{0000-0002-8175-874X}

\author{Xinpeng Li}
\affiliation{%
	\department{School of Computer Science and Technology}
	\institution{East China Normal University}
	\city{Shanghai}
	\country{China}}
\email{51275901118@stu.ecnu.edu.cn}
\orcid{0009-0002-4103-5317}

\author{Xi Mao}
\affiliation{%
	\department{School of Design}
	\institution{East China Normal University}
	\city{Shanghai}
	\country{China}}
\email{xmao@design.ecnu.edu.cn}
\orcid{0009-0004-0358-7780}

\author{Changbo Wang}
\affiliation{%
	\department{School of Computer Science and Technology}
	\institution{East China Normal University}
	\city{Shanghai}
	\country{China}}
\email{cbwang@cs.ecnu.edu.cn}
\orcid{0000-0001-8940-6418}

\author{Chenhui Li}
\authornote{Corresponding author.}
\affiliation{%
	\department{School of Computer Science and Technology}
	\institution{East China Normal University}
	\city{Shanghai}
	\country{China}}
\email{chli@cs.ecnu.edu.cn}
\orcid{0000-0001-9835-2650}


\begin{abstract}
Artistic image assessment has become a prominent research area in computer vision. In recent years, the field has witnessed a proliferation of datasets and methods designed to evaluate the aesthetic quality of paintings. However, most existing approaches focus solely on static final images, overlooking the dynamic and multi-stage nature of the artistic painting process. To address this gap, we propose a novel framework for human-aligned assessment of painting processes. Specifically, we introduce the Painting Process Assessment Dataset (PPAD)—the first large-scale dataset comprising real and synthetic painting process images, annotated by domain experts across eight detailed attributes. Furthermore, we present PPJudge (Painting Process Judge), a Transformer-based model enhanced with temporally-aware positional encoding and a heterogeneous mixture-of-experts architecture, enabling effective assessment of the painting process. Experimental results demonstrate that our method outperforms existing baselines in accuracy, robustness, and alignment with human judgment, offering new insights into computational creativity and art education.
\end{abstract}





\maketitle

\section{Introduction}
In recent years, with the rapid development of computer vision and generative models, intelligent assessment has shown great potential in the field of art. In painting-related research, existing painting aesthetic assessment methods~\cite{DBLP:conf/aaai/JiangLSGWL24,DBLP:conf/cvpr/YiTGLR23,DBLP:conf/ijcai/0015QLWGHL24} are capable of evaluating static artworks from dimensions such as composition, color, and style, thereby advancing computational understanding of visual aesthetics. However, in the context of art education, painting is regarded as a multi-stage dynamic practice that involves observation, ideation, composition, and execution~\cite{Botella2016,Botella2022}. Therefore, the assessment of the painting process is particularly crucial, as it more accurately reflects the painter’s abilities in observation, reasoning, and execution. By analyzing quantitative attributes such as stability and reasoning depth~\cite{Botella2018}, researchers can obtain a more comprehensive understanding of a painter’s artistic ability.

Traditionally, the assessment of the painting process has been conducted by experts in art or design, who evaluate factors such as consistency, stability, and reasoning depth throughout the process to infer the artist's level. However, this approach is inherently time-consuming and may be influenced by personal biases or preconceived notions. With the recent development of generative models, researchers have begun reconstructing human painting processes using diffusion-based approaches~\cite{DBLP:conf/cvpr/RombachBLEO22}. While these methods~\cite{DBLP:conf/siggrapha/SongHYCYLZS24,DBLP:conf/iccv/LiuLHLDLDW21} can serve as useful references for understanding the painting process, they cannot be regarded as definitive standards. This is because their training objectives focus primarily on consistency with the final outcome, without any direct supervision of the intermediate steps. As a result, the generated painting processes often follow fixed patterns and lack the diversity and creativity characteristic of genuine human creation.

Recently, researchers have employed multimodal models to tackle tasks related to painting assessment~\cite{DBLP:conf/mm/BinSDH00NS24,DBLP:conf/nips/JinQLWHGLL24}. However, these approaches face two major challenges when applied to painting process assessment. First, there is currently no annotated dataset specifically designed for this task. Existing datasets~\cite{DBLP:conf/aaai/JiangLSGWL24,DBLP:conf/nips/JinQLWHGLL24} focus on evaluating the content of the final image, rather than the painting process. Second, current assessment models typically operate on a single image and are not designed to handle multi-image inputs that represent a sequence of painting steps.

To overcome existing challenges, we introduce a dataset specifically designed for painting process assessment: the Painting Process Assessment Dataset (PPAD). It consists of approximately 15,000 real paintings and 10,000 synthetic paintings, each annotated by domain experts. To the best of our knowledge, PPAD is the first dataset dedicated to studying and evaluating the painting process. In addition, we propose PPJudge (Painting Process Judge), which effectively leverages both category-level and process-level features of a given painting. PPJudge is a significant extension of the Transformer architecture with Mixture-of-Experts (MoE) modules. First, we introduce an angular correction mechanism in positional encoding, enabling the model to better capture the sequential nature of painting. Second, we employ a heterogeneous MoE architecture, where experts differ in depth, allowing the model to adaptively select appropriate experts based on different levels of attributes in the painting process. Furthermore, we constrain the shared expert modules by painting categories, ensuring the evaluation is context-aware and category-sensitive.

In summary, our main contributions are as follows:
\begin{itemize}
	\item We address the underexplored problem of painting process assessment and introduce a new dataset, PPAD, consisting of about 15,000 real and 10,000 synthetic paintings, each with expert annotations.
	
	\item We propose a novel painting process assessment model called PPJudge, which incorporates temporal-aware position encoding and a heterogeneous MoE architecture.
	
	\item Extensive experiments on PPAD demonstrate the superior performance of our model in assessing the quality of painting processes.
\end{itemize}

\section{Related Work}

\subsection{Painting Assessment Dataset}
In earlier years, \cite{DBLP:conf/mm/YanulevskayaUBSZBMS12} annotated 500 abstract paintings by rating them on a Likert scale of 1-7 to assess emotional valence. To explore the impact of subjectivity, JenAesthetics~\cite{DBLP:conf/eccv/AmirshahiHDR14} includes 1,628 oil paintings along with results from a subjective study. Furthermore, JenAesthetics$\beta$\cite{DBLP:journals/corr/AmirshahiHDR16} extends this dataset by adding 281 high-quality paintings and analyzing various factors influencing painting assessment. Building upon these, VAPS\cite{VAPS2022} comprises 999 paintings spanning five genre categories. WikiArt~\footnote{\url{https://www.wikiart.org/}} is a widely used painting dataset that contains 81,449 artworks, annotated with 27 styles and 45 genres. Similar to WikiArt, Art500k~\cite{DBLP:conf/mm/MaoCS17} includes over 500K artistic images, each labeled with more than 10 attributes. OmniArt~\cite{DBLP:journals/tomccap/StrezoskiW18} is a large-scale dataset with over one million artworks, providing a rich annotation structure. SemArt~\cite{DBLP:conf/eccv/GarciaV18} is a multi-modal dataset for semantic art understanding, containing 21,383 paintings with detailed metadata. More recently, BAID~\cite{DBLP:conf/cvpr/YiTGLR23} is introduced, comprising 60,337 artistic images across various art forms, with more than 360,000 user-contributed votes for aesthetic evaluation. AACP dataset~\cite{DBLP:conf/aaai/JiangLSGWL24} contains 21,200 children’s paintings, including 20,000 unlabeled paintings generated by DALL-E~\cite{DBLP:journals/corr/abs-2204-06125} and 1,200 expert-annotated paintings.

Unlike the painting datasets mentioned above, the dataset we constructed is specifically designed to capture the painting process. It contains both real and AI-generated data, all of which have been expertly labeled with multiple attributes.

\subsection{Painting Assessment Model}
Unlike image aesthetics assessment (IAA) models, which have developed rapidly thanks to large-scale IAA datasets, painting assessment models have progressed at a slower pace. Early painting assessment models focused on classifying paintings into styles or genres~\cite{4545847,DBLP:conf/eccv/CarneiroSBC12,DBLP:journals/tap/ShamirMOEG10,DBLP:conf/icip/TanCAT16}. With the advancement of multimodal models, many studies have started to explore painting evaluation using multimodal approaches. CLIP-Art~\cite{DBLP:conf/cvpr/CondeT21} utilizes CLIP~\cite{DBLP:conf/icml/RadfordKHRGASAM21} to encode paintings into embeddings for downstream tasks. Additionally, SAAN~\cite{DBLP:conf/cvpr/YiTGLR23} introduces style-specific and generic aesthetic features via spatial information fusion for painting assessment. AACP~\cite{DBLP:conf/aaai/JiangLSGWL24} introduces synthetic paintings through a self-supervised learning strategy to address the challenge of insufficient painting annotations, marking the first use of synthetic data in painting evaluation. Recently, GalleryGPT~\cite{DBLP:conf/mm/BinSDH00NS24} fine-tunes a multimodal large language model to generate formal analyses for paintings. ArtCLIP~\cite{DBLP:conf/nips/JinQLWHGLL24} proposes a multi-attribute contrastive learning framework for painting assessment across multiple categories and attributes.

In this paper, we aim to evaluate the painting process rather than the final painting. Unlike previous methods that map a painting to one or more scores, our method processes multiple paintings simultaneously and analyzes differences across different stages of the painting process.

\begin{figure*}[]
	\centering
	\includegraphics[width=1\linewidth]{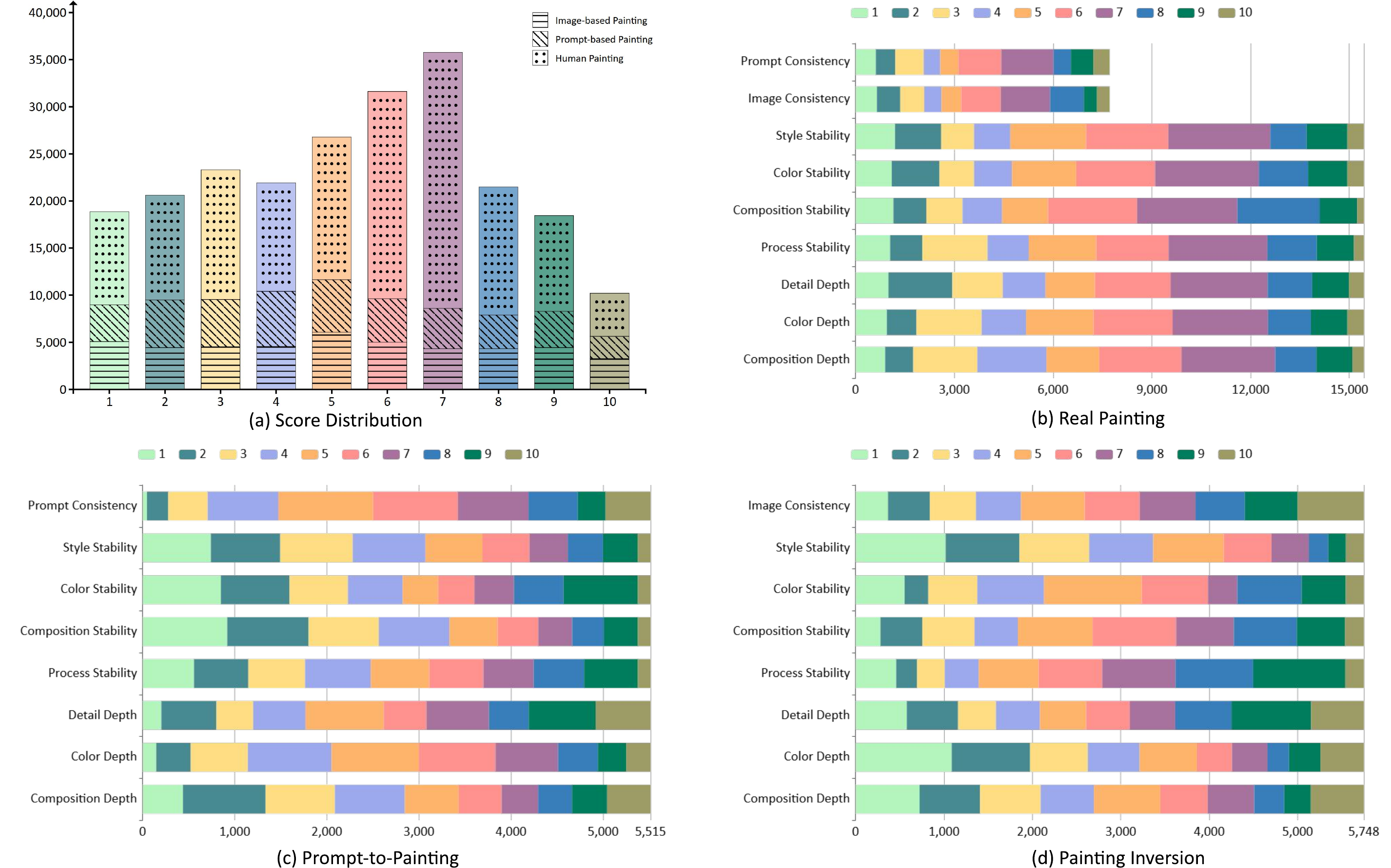}
	\caption{Statistics of our dataset. (a) shows the distribution of scores for paintings from different sources. (b), (c), and (d) show the distribution of painting attributes from real paintings, prompt-to-painting generation, and painting inversion, respectively.}
	\label{3}
\end{figure*}

\subsection{Sequence Image Representation}
Sequence image representation has evolved significantly over the past decades, covering both image and video domains. Convolutional neural networks (CNNs) were long the dominant architecture for visual representation learning~\cite{DBLP:conf/cvpr/HeZRS16,DBLP:conf/cvpr/HuangLMW17}. Recently, vision transformers (ViTs) have become popular image encoders. DeiT~\cite{DBLP:conf/icml/TouvronCDMSJ21} introduces knowledge distillation to enable data-efficient ViT training, while Swin Transformer~\cite{DBLP:conf/iccv/LiuL00W0LG21} proposes hierarchical attention with local windows for improved efficiency and scalability.
To further enhance image encoders, researchers explore sparse and lightweight architectures~\cite{DBLP:journals/jmlr/FedusZS22}.

Compared to image encoders, video encoders additionally capture temporal dynamics across frames. Early approaches extended 2D CNNs to 3D to model spatial and temporal features~\cite{DBLP:conf/cvpr/CarreiraZ17}. Recently, many works have adapted Transformers to video representation. ViViT~\cite{DBLP:conf/iccv/Arnab0H0LS21} proposes factorized attention strategies for efficient spatio-temporal modeling, while TimeSformer~\cite{DBLP:conf/icml/BertasiusWT21} decouples spatial and temporal attention to reduce complexity. Video Swin Transformer~\cite{DBLP:conf/cvpr/LiuN0W00022} extends hierarchical attention to the temporal dimension.
Self-supervised video pretraining has also gained attention. VideoMAE~\cite{DBLP:conf/nips/TongS0022,DBLP:conf/cvpr/WangHZTHWWQ23} utilizes masked video modeling to learn strong temporal-spatial representations from unlabeled videos.

Image encoders process sequential images by simply aggregating the embeddings of individual frames, which fails to effectively capture inter-frame relationships. In contrast, video encoders model temporal dynamics but require re-encoding the entire frame sequence at each step. In this paper, we propose an efficient design based on an image encoder augmented with key-value caching. Our method supports incremental updates, avoiding redundant computation and significantly improving inference efficiency.

\section{Dataset}

Our Painting Process Assessment Dataset (PPAD) consists of pairs $(R_m, P_m)$, where $R_m$ denotes the $m$-th reference painting or prompt, and $P_m = \{P_{m1}, P_{m2}, ..., P_{mn}\}$ represents a sequence of $n$ paintings corresponding to $R_m$. We first collect a large number of real-world painting process images (Section~\ref{3.1}). To improve the generalization ability of models, we further introduce synthetic painting process images (Section~\ref{3.2}). Finally, we invite domain experts to evaluate the painting processes using eight attributes (Section~\ref{3.3}).

\subsection{Painting Collection}
\label{3.1}
Following~\cite{DBLP:conf/aaai/JiangLSGWL24} and~\cite{DBLP:conf/chi/LiuBYCJTGWL24}, we recruit hundreds of art and drawing majors as participants aged between 22 and 30. Each participant is given either a random prompt or a reference painting. Participants can create freely based on the prompt or replicate the reference faithfully. The entire painting process is recorded as a canvas-focused video. Since participants often pause to observe or think, videos contain redundant frames, and painting durations vary widely. To address this, we sample one frame every $k$ frames and compute visual differences between consecutive frames. A frame is selected as a key frame if its visual difference from the last selected key frame exceeds a threshold $\tau$, which is dynamically adjusted based on the total visual change across the video. To ensure consistent representation, we further constrain the number of key frames within predefined bounds.

After filtering out invalid and incomplete paintings, we collected approximately 15,000 paintings, 80\% of which are allocated for training and the remaining 20\% for testing.

\begin{figure*}[]
	\centering
	\includegraphics[width=1\linewidth]{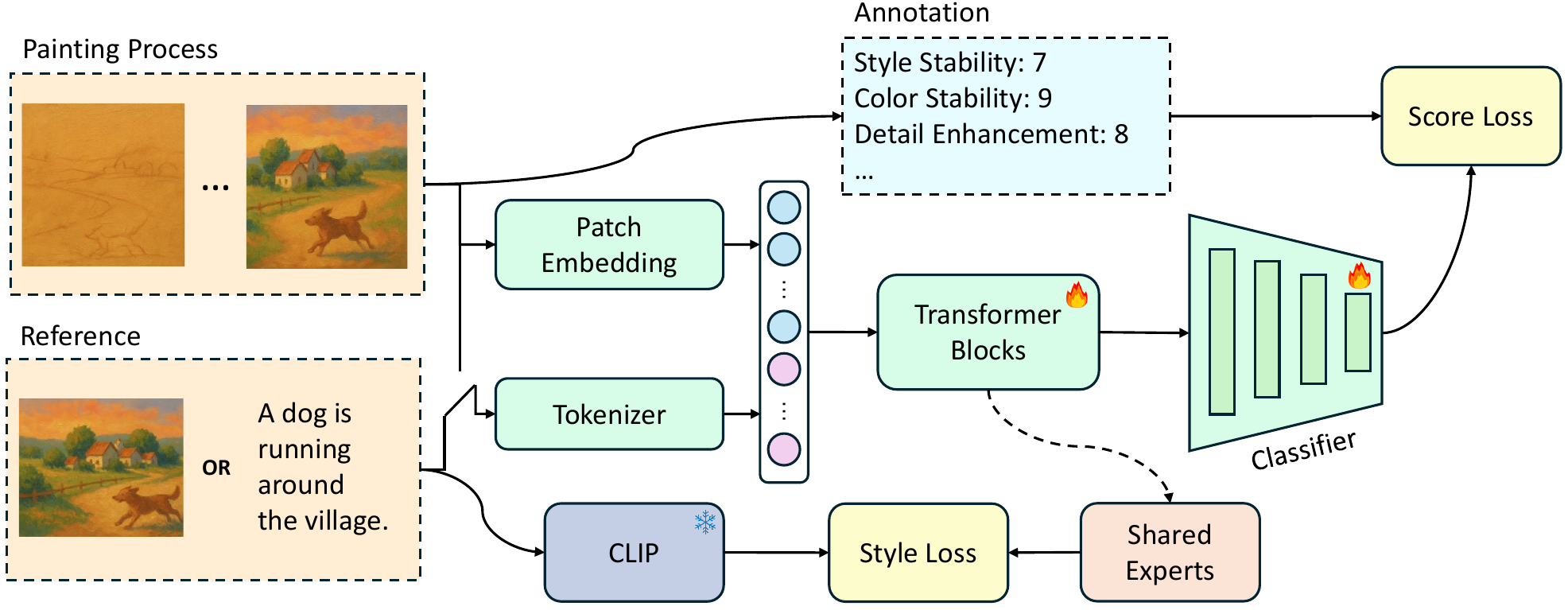}
	\caption{Pipeline of our model. Process images and a reference (either an image or a prompt) are first embedded into token representations. Improved transformer blocks extract features of the painting process, with a style loss guiding the shared experts toward capturing stylistic characteristics. The extracted features are then mapped to assessment scores by a classifier. The transformer blocks and the classifier are jointly optimized with a score loss.}
	\label{overview}
\end{figure*}

\subsection{Dataset Expansion}
\label{3.2}
To improve the generalization ability of the model, we introduce synthetic data. Early painting reconstruction efforts, such as~\cite{huang2019learning}, were inspired by human painters and trained using reinforcement learning. Later, \cite{DBLP:conf/cvpr/ZouSQ0S21} and~\cite{DBLP:conf/iccv/LiuLHLDLDW21} used a stroke-based rendering approach to improve the stability of painting reconstruction. However, the paintings reconstructed by such methods exhibit significant differences from the target paintings. Recently, diffusion-based models~\cite{DBLP:conf/cvpr/RombachBLEO22} have achieved great success. Therefore, \cite{DBLP:conf/siggrapha/0003WCKS24,DBLP:conf/siggrapha/SongHYCYLZS24,paintsundo} have leveraged diffusion models to generate high-quality process images.

In this paper, we obtain synthetic data in two ways. First, we use ProcessPainter~\cite{DBLP:conf/siggrapha/SongHYCYLZS24} to generate about 5,000 paintings along with process images from prompts. These prompts are simple descriptions of themes or scenes, with no painting style specified. Second, we use Paints-Undo~\cite{paintsundo} to invert real paintings to obtain about 5,000 process images. 90\% of the synthetic data is used to pre-train our models, while the remaining 10\% is used for testing.

\subsection{Painting Annotation}
\label{3.3}
Fifteen experts in the field of art are invited to evaluate the paintings. Based on fundamental drawing principles and design theory, the experts divided painting process assessment into three aspects: \textit{Painting Consistency}, \textit{Painting Stability}, and \textit{Painting Depth}.

\textbf{Painting Consistency} (Consis.) evaluates whether the final painting accurately reflects the core themes and elements of the given prompt or reference image, maintaining strong semantic and visual similarity.

\textbf{Painting Stability} evaluates the smoothness and coherence of transitions during the painting process. This attribute is assessed across four dimensions: \textit{style}, \textit{color}, \textit{composition}, and \textit{process}. Style stability (S. S.) refers to whether the sequence maintains a consistent artistic style without abrupt or inconsistent changes; Color stability (Col. S.) assesses whether the color transitions are gradual and harmonious, avoiding sudden or jarring shifts such as the appearance of unnatural hues; Composition stability (Com. S.) considers whether the visual focus or layout changes drastically across frames; Process stability (Proc. S.) evaluates whether the painting progresses smoothly without excessive corrections, regressions, or unnecessary modifications.

\textbf{Painting Depth} assesses whether the painting process exhibits progressive refinement through the gradual introduction of new techniques or complex visual elements. This aspect is evaluated in terms of \textit{detail}, \textit{color}, and \textit{composition}. Detail depth (D. D.) measures the incremental enrichment of visual details and their contribution to expressive power; Color depth (Col. D.) evaluates whether the use of color becomes more sophisticated over time, enhancing visual impact; Composition depth (Com. D.) assesses whether compositional structures grow in complexity and contribute to the overall visual richness.

In total, the painting process is assessed using eight attributes on a scale from 1 to 10. Specifically, each painting’s drawing process is evaluated by multiple experts, and each attribute receives multiple scores. The final score for each attribute is computed as the average of these scores.

\section{Method}
\begin{figure*}[]
	\centering
	\includegraphics[width=1\linewidth]{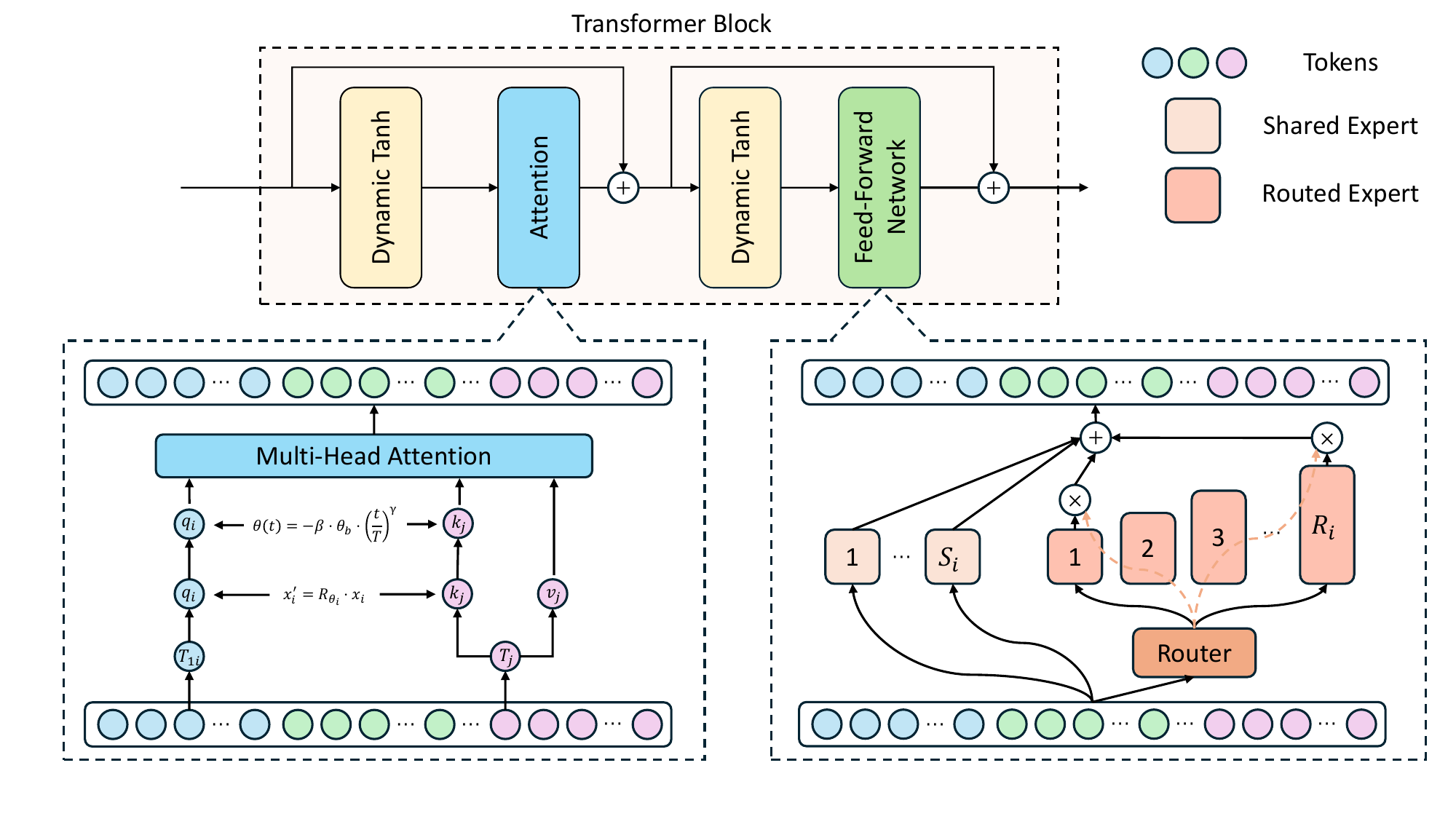}
	\caption{Illustration of the basic architecture of our transformer blocks.}
	\label{4}
\end{figure*}
\subsection{Overall Architecture}
The assessment of the painting process can be represented as $Scores = M(P_m | R_m)$, where $M$ denotes the model. Our model consists of transformer blocks and a classifier. The overall structure of the transformer block is shown in~\autoref{4}. To adapt the model to 2D patches extracted from sequential images, we apply an angular adjustment to the original rotary positional encoding (Section~\ref{Section: 4.2}). Next, to allow the model to adapt to different types of paintings and better disentangle painting features, we propose a heterogeneous mixture of experts (Section~\ref{Section: 4.3}). Finally, we describe our classifier and the loss function in Section~\ref{Section: 4.4}.

\subsection{Positional Encoding}
\label{Section: 4.2}
Rotary Position Embedding (RoPE)~\cite{DBLP:journals/ijon/SuALPBL24} is widely used in multimodal large language models~\cite{Qwen2.5-VL,DBLP:journals/corr/abs-2407-21783}. RoPE improves generalization by mapping token representations into a high-dimensional rotational space, which can be formalized as:
\begin{equation}
	\alpha_{ij}^{RoPE} = q_{i}R(i-j)k_{j}^{T},
\end{equation}
where $ q_i = W_Q(x_i) $ is the query vector from the $ i $-th token, and $ k_j = W_K(x_j) $ is the key vector from the $ j $-th token. $ R(i-j) $ is a block diagonal matrix that applies a position-dependent rotation to the query and key vectors.

However, this approach is not well-suited for sequential image modeling, where spatial relationships between tokens should be preserved. Given an \( n \)-step sequence of paintings \( P = \{p_1, ... , p_n\} \), it is tokenized into \( T = \{\{t_{11}, ..., t_{1m}\}, ..., \{t_{n1}, ... , t_{nm}\}\} \). Although \( t_{21} \) and \( t_{11} \) occupy the same spatial location in adjacent frames, their positions in the flattened token sequence are far apart, resulting in exaggerated angular separation under standard RoPE.

To correct for this discrepancy, we introduce a time-dependent correction term, with later frames receiving more negative corrections, and earlier frames having values closer to zero to mitigate temporal angular drift. Formally, given the standard RoPE rotation matrix parameterized by position $p$, we introduce a temporal adjustment to the rotation angle as:
\begin{equation}
	\theta_{\text{time}}(t) = -\beta \cdot \theta_{\text{base}} \cdot \left( \frac{t}{T} \right)^\gamma,
\end{equation}
where $t$ is the time step, $T$ is the total number of time steps, $\theta_{\text{base}} = \theta_{\text{RoPE}}(p=0)$ is the baseline rotation angle at the first spatial token, $\beta$ is a tunable factor controlling the degree of temporal correction, and $\gamma$ is an optional smoothing exponent that adjusts the rate of temporal decay.

\begin{table*}[]
	\centering
	\caption{Comparison of AANSPS, SAAN, ArtCLIP and ours on PPAD.}
	\begin{tabular}{l|cccc|cccc|cccc|cccc}
		\toprule[1.5pt]
		\multirow{2}{*}{Attribute} & \multicolumn{4}{c|}{AACP} & \multicolumn{4}{c|}{SAAN} & \multicolumn{4}{c|}{ArtCLIP}  & \multicolumn{4}{c}{\textbf{Ours}} \\
		& SRCC  & PCC   & MSE   & ACC   & SRCC  & PCC  & MSE    & ACC   & SRCC   & PCC   & MSE   & ACC   & SRCC   & PCC   & MSE       & ACC    \\ \midrule
		Consis.      & 0.58  & 0.60  & 0.24  & 0.84  & 0.68  & 0.68 & 0.40   & 0.75  & 0.76   & 0.75  & 0.37  & 0.71  & \textbf{0.78}   & \textbf{0.79}  & \textbf{0.12}      &\textbf{0.92}           \\
		S. S.        & 0.49  & 0.47  & 0.42  & 0.73  & 0.65  & 0.61 & 0.53   & 0.79  & 0.67   & 0.59  & 0.35  & 0.83  & \textbf{0.66}   & \textbf{0.65}  & \textbf{0.22}      &\textbf{0.85}           \\
		Col. S.      & 0.62  & 0.61  & 0.22  & 0.81  & 0.64  & 0.71 & 0.38   & 0.81  & 0.65   & 0.70  & 0.46  & 0.58  & \textbf{0.75}   & \textbf{0.78}  & \textbf{0.11}      &\textbf{0.89}          \\
		Com. S.      & 0.64  & 0.63  & 0.23  & 0.82  & 0.64  & 0.68 & 0.33   & 0.75  & 0.52   & 0.45  & 0.52  & 0.88  & \textbf{0.77}   & \textbf{0.75} & \textbf{0.16}      &\textbf{0.90}          \\
		Proc. S.     & 0.60  & 0.64  & 0.30  & 0.72  & 0.69  & 0.68 & 0.45   & 0.75  & 0.63   & 0.64  & 0.35  & 0.79  & \textbf{0.70}   & \textbf{0.73}  & \textbf{0.16}      &\textbf{0.85}          \\
		D. D.        & 0.59  & 0.61  & 0.34  & 0.69  & 0.47  & 0.49 & 0.65   & 0.63  & 0.69   & 0.76  & 0.27  & 0.75  & \textbf{0.74}   & \textbf{0.73}  & \textbf{0.14}      &\textbf{0.87}           \\
		Col. D.      & 0.62  & 0.65  & 0.30  & 0.73  & 0.67  & 0.66 & 0.26   & 0.78  & 0.60   & 0.59  & 0.18  & 0.75  & \textbf{0.69}   & \textbf{0.68}  & \textbf{0.20}      &\textbf{0.86}          \\
		Com. D.      & 0.59  & 0.56  & 0.29  & 0.71  & 0.47  & 0.49 & 0.70   & 0.67  & 0.66   & 0.72  & 0.31  & 0.71  & \textbf{0.80}   & \textbf{0.76}  & \textbf{0.13}      &\textbf{0.91}     \\ \bottomrule[1.5pt]
	\end{tabular}
	\label{5.3}
\end{table*}

\subsection{Heterogeneous Mixture of Experts}
\label{Section: 4.3}
MoE offers advantages in terms of faster convergence and improved computational efficiency by leveraging sparse activation. In multi-attribute prediction tasks, using a single FFN often results in feature entanglement~\cite{DBLP:conf/aaai/JiangLSGWL24}, making the learning process slower and reducing generalization. MoE mitigates this issue by dynamically selecting specialized experts, thereby enhancing both learning efficiency and model robustness.

In painting process assessment, different visual attributes require features at varying levels of abstraction. For example, color-related attributes often depend on shallow features, while composition-related attributes require deeper, more global representations. To accommodate these heterogeneous needs, we design a heterogeneous MoE architecture, as shown in~\autoref{4}. Similar to ~\cite{DBLP:journals/corr/abs-2412-19437}, our MoE consists of two types of experts: shared and routed. The shared experts are conditioned on the painting style, which plays an essential role in shaping the painting process. We leverage CLIP~\cite{DBLP:conf/icml/RadfordKHRGASAM21} embeddings to provide semantic supervision for style-aware shared experts. The routed experts are selected dynamically by a router and exhibit varying depths, allowing them to specialize in different attribute types. This design enables the model to process diverse artistic attributes in a more targeted and efficient manner.

\subsection{Classifier and Loss Function}
\label{Section: 4.4}
The classifier maps the model output to the scores. Unlike AACP, our backbone already has the ability to decouple features. Following Laion’s aesthetic score predictor~\cite{DBLP:conf/nips/SchuhmannBVGWCC22}, we use a multi-layer MLP as the classifier. 

During training, we use two loss functions: style loss and score loss. The style loss is meant to constrain the sharing expert to learn style-specific features. First, we use a pre-trained CLIP model to distinguish between different types of paintings, which can be expressed as:
\begin{equation}
	t^* = \arg\max_{t \in T} \frac{E_p \cdot t}{\|E_p\| \|t\|},
\end{equation}
where \( T \) represents a set of text embeddings, where each text is of the form "This is a \{style\} painting.", and \( E_p \) denotes the reference image embedding. We set \( E_{style} = t^* \). Second, the output of the shared experts is projected into the style dimension using a projection module, which can be expressed as:
\begin{equation}
	E_{se}^{l} = W_{proj} \cdot O_{se}^{l},
\end{equation}
where \( W_{proj} \) is the learnable projection matrix and \( O_{se}^{l} \) represents the output of the shared expert at layer \( l \). We use cosine similarity loss to calculate the style loss, which can be expressed as:
\begin{equation}
	L_{style}^{l} = 1 - \frac{E_{style} \cdot E_{se}^{l}}{\|E_{style}\| \|E_{se}^{l}\|}.
\end{equation}
Finally, the total style loss is defined as:
\begin{equation}
	L_{style} = \sum_{l=1}^{L} \alpha_l \cdot L_{style}^{l},
\end{equation}
where \( \alpha_l \) is the weighting coefficient for layer \( l \).

The score loss is calculated using the Mean Squared Error (MSE), which can be expressed as:
\begin{equation}
	L_{score} = \frac{1}{n} \sum_{i=1}^{n} ( y_{score}^{(i)} - \hat{y}_{score}^{(i)} )^2,
\end{equation}
where \( n \) denotes the number of attributes.

Therefore, the total loss can be expressed as:
\begin{equation}
	L_{total} =  L_{style} + \lambda_{score} L_{score},
\end{equation}
where \( \lambda_{score} \) is a weighting coefficient to balance the style loss and score loss.

\section{Experiment}

\subsection{Experiment Setting}
We set the number of Transformer blocks to 8. The  classifier consists of a 3-layer MLP and uses the SiLU activation function. We set the hyperparameter \( \alpha \) in the loss function to increase with \( l \) from 0 to 0.8, following an exponential schedule. The weighting coefficient \( \lambda_{score} \) is set to 10. Our model is trained on 8 NVIDIA H800 GPUs with a batch size of 256. During the pre-training phase with synthetic data, we train the model for 200 epochs using the AdamW optimizer with a learning rate of $1 \times 10^{-4}$. In the subsequent training phase with real data, we employ the Adam optimizer with a learning rate of $1 \times 10^{-5}$ and train for 20 epochs.

\subsection{Metrics}
To evaluate the performance of each model, we adopt three widely used metrics: 1) Spearman’s Rank Correlation Coefficient (SRCC), 2) Pearson Correlation Coefficient (PCC), and 3) Mean Squared Error (MSE). Additionally, we compute classification accuracy (ACC) by rounding both the predicted scores to the nearest integer and considering a prediction correct if the rounded values match.

\begin{figure*}[]
	\centering
	\includegraphics[width=1\linewidth]{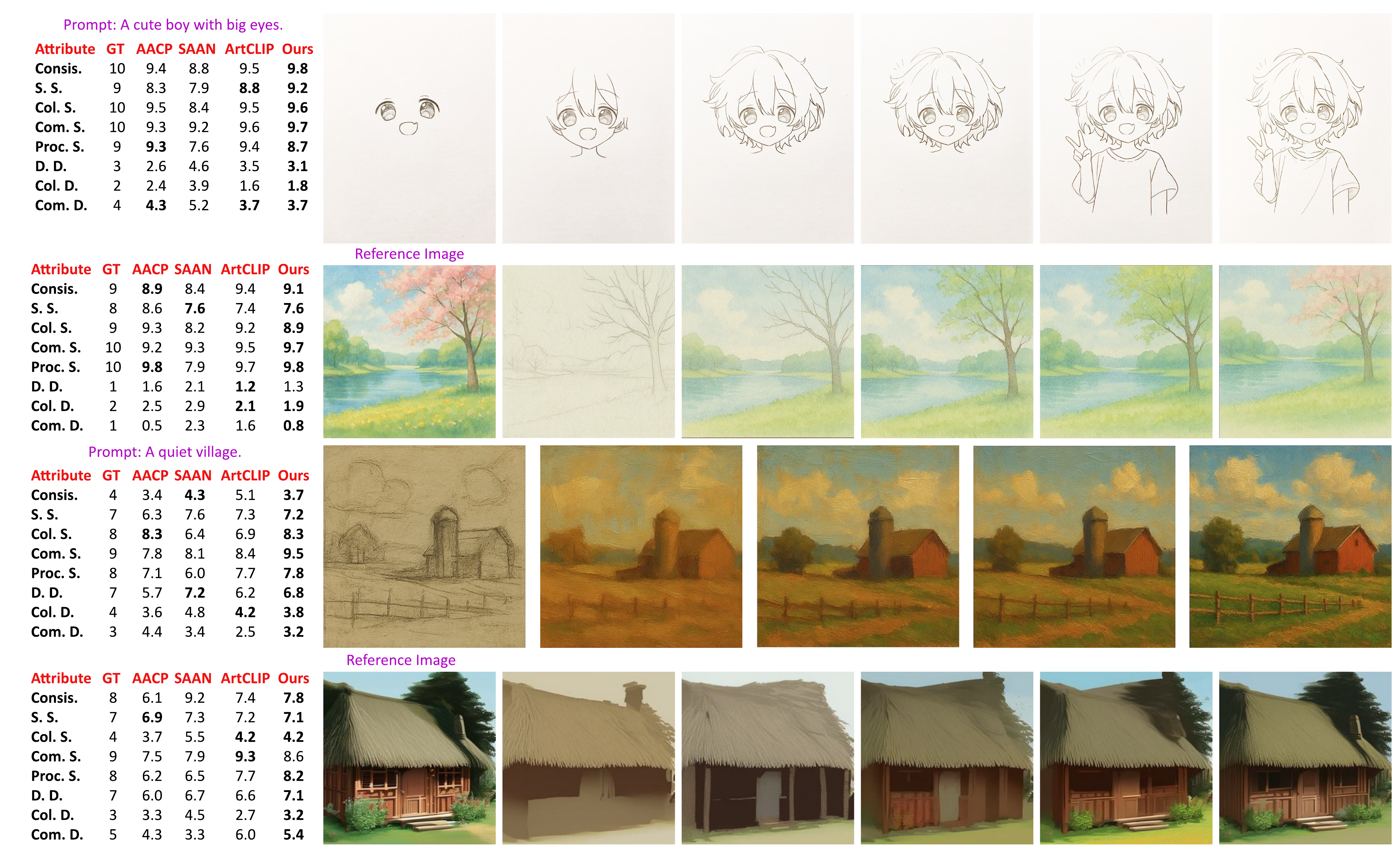}
	\caption{Examples from our testset. The top two rows are real paintings and the bottom two rows are synthesized paintings. The painting is based on Prompt or Reference image. We show the scores labeled by the experts as well as the scores predicted by painting assessment models.}
	\label{figure:5.3}
\end{figure*}

\subsection{Quantitative Experiment}
We compare our method with several advanced painting assessment models, including AACP~\cite{DBLP:conf/aaai/JiangLSGWL24}, SAAN~\cite{DBLP:conf/cvpr/YiTGLR23}, and ArtCLIP~\cite{DBLP:conf/nips/JinQLWHGLL24}. All models are retrained on our PPAD dataset using their recommended configurations. Since AACP, SAAN, and ArtCLIP are designed for single-image input and do not support multi-image sequences, we adapt them by averaging the embeddings of multiple images and feeding the aggregated feature into the classifier. 

\autoref{figure:5.3} shows examples of our dataset and comparison results. From the prediction scores presented, our method is significantly more accurate than the other methods. As shown in~\autoref{5.3}, our method outperforms all baselines across all metrics and attributes, demonstrating superior capability in assessing the painting process.

Furthermore, we compare the inference performance of each model. All experiments are conducted on a single NVIDIA H800 GPU, with the number of input images for each painting process set to 10. As shown in~\autoref{table: 5.3.2}, our method achieves the fastest inference time among all methods. This demonstrates the potential of our approach to provide real-time feedback to the artist during the painting process.

\begin{table}[]
	\centering
	\caption{Comparison of inference performance.}
	\begin{tabular}{lcc}
		\toprule[1.5pt]
		Model   & Parameter (M) & Time (s) \\ \midrule
		AACP    & 62.17         & 0.03     \\
		SAAN    & 30.82         & 0.06     \\
		ArtCLIP & 149.62        & 0.04     \\ \midrule
		\textbf{Ours}    & \textbf{17.21}         & \textbf{0.01}      \\ \bottomrule[1.5pt]
	\end{tabular}
	\label{table: 5.3.2}
\end{table}

\subsection{Ablation Study}
As shown in~\autoref{5.3.1}, we start with the base Transformer equipped with a standard MoE architecture and incrementally introduce new modules. The performance steadily improves with the addition of angular corrections in positional encoding, variable-length expert modules, and style loss.

\begin{table}[]
	\centering
	\caption{Ablation studies on model components.}
	\begin{tabular}{lcccc}
		\toprule[1.5pt]
		Model                      & SRCC     & LCC   & MSE   & ACC\\ \midrule
		Base                       & 0.48     & 0.43  & 0.34  & 0.61\\
		~~ + Angular Corrections   & 0.59     & 0.61  & 0.29  & 0.68\\
		~~ + Variable length       & 0.67     & 0.66  & 0.19  & 0.82\\
		~~ + Style Loss            & 0.74     &0.73   &0.16   & 0.88 \\ 		\bottomrule[1.5pt]
	\end{tabular}
	\label{5.3.1}
\end{table}

Furthermore, we investigate the impact of the number of shared experts and the number of activated routing experts per token. In our experimental setup, each token has a hidden dimension of 512, and the total number of experts is set to 32. As shown in~\autoref{5.3.3}, we conduct five groups of ablation experiments and find that the model achieves near-optimal performance when using 2 shared experts and 4 routing experts. Increasing the number of shared or routing experts beyond this point yields only marginal improvements.

Based on the previous results, we select the optimal expert configuration and further investigate the impact of expert depth heterogeneity under a fixed number of activated routed experts. As shown in \autoref{5.3.4}, we compare four configurations: (1) A1, where all experts are 1-layer MLPs; (2) A2, where all experts are 2-layer MLPs; (3) A3, which includes six experts with varying depths from 1 to 5; and (4) Ours, a structured heterogeneous setting with 16 experts of depth 1, 6 of depth 2, 4 of depth 3, and 2 each of depths 4 and 5. Our configuration consistently achieves the best performance across all four evaluation metrics, demonstrating the effectiveness of incorporating structured depth diversity among routed experts.

\begin{table}[]
	\centering
	\caption{Ablation study on the number of (Shared experts + Routed experts).}
	\begin{tabular}{lcccc}
		\toprule[1.5pt]
		Nums.     & SRCC  & LCC     & MSE     & ACC \\ \midrule
		1 + 2    & 0.65  & 0.63    & 0.27    & 0.80    \\
		2 + 2    & 0.71  & 0.69    & 0.21    & 0.85    \\
		2 + 4    & \textbf{0.74}  & 0.73    & \textbf{0.16}    & 0.88     \\ \midrule
		2 + 6    & 0.73  & 0.73    & 0.17    & \textbf{0.89}\\
		4 + 4    & \textbf{0.74}  & \textbf{0.75}    & 0.17    & \textbf{0.89}   \\ \bottomrule[1.5pt]
	\end{tabular}
	\label{5.3.3}
\end{table}

\begin{table}[]
	\centering
	\caption{Comparison of expert depth configurations with fixed number of activated routed experts (Top-4).}
	\label{tab:expert-depth-ablation}
	\begin{tabular}{lcccc}
		\toprule[1.5pt]
		Expert Depth  & SRCC  & LCC  & MSE  & ACC  \\
		\midrule
		A1            & 0.69  & 0.68  & 0.22  & 0.83     \\
		A2            & 0.71  & 0.68  & 0.23  & 0.82     \\
		A3            & 0.62  & 0.60  & 0.31  & 0.76     \\ \midrule
		\textbf{Ours}          & \textbf{0.74}  & \textbf{0.73}  & \textbf{0.16}  & \textbf{0.88}     \\
		\bottomrule[1.5pt]
	\end{tabular}
	\label{5.3.4}
\end{table}

In addition, we analyze the impact of synthetic data on model performance. As shown in~\autoref{5.3.2}, the model achieves better results when synthetic data is progressively incorporated alongside real data, demonstrating the benefit of data augmentation in enhancing model generalization.

\begin{table}[]
	\centering
	\caption{Effect of Synthetic Data on Model Performance.}
	\begin{tabular}{lcccc}
		\toprule[1.5pt]
		Data            	    	 & SRCC    & LCC   & MSE  & ACC\\ \midrule
		Real data             		 & 0.68    & 0.66  & 0.23 & 0.84\\
		~~ + 50\% Synthetic data     & 0.71    & 0.69  & 0.19 & 0.85\\
		~~ + 50\% Synthetic data     & 0.74    & 0.72  & 0.16 & 0.88 \\		
		\bottomrule[1.5pt]
	\end{tabular}
	\label{5.3.2}
\end{table}

\subsection{User Study}
\begin{figure}[]
	\centering
	\includegraphics[width=1\linewidth]{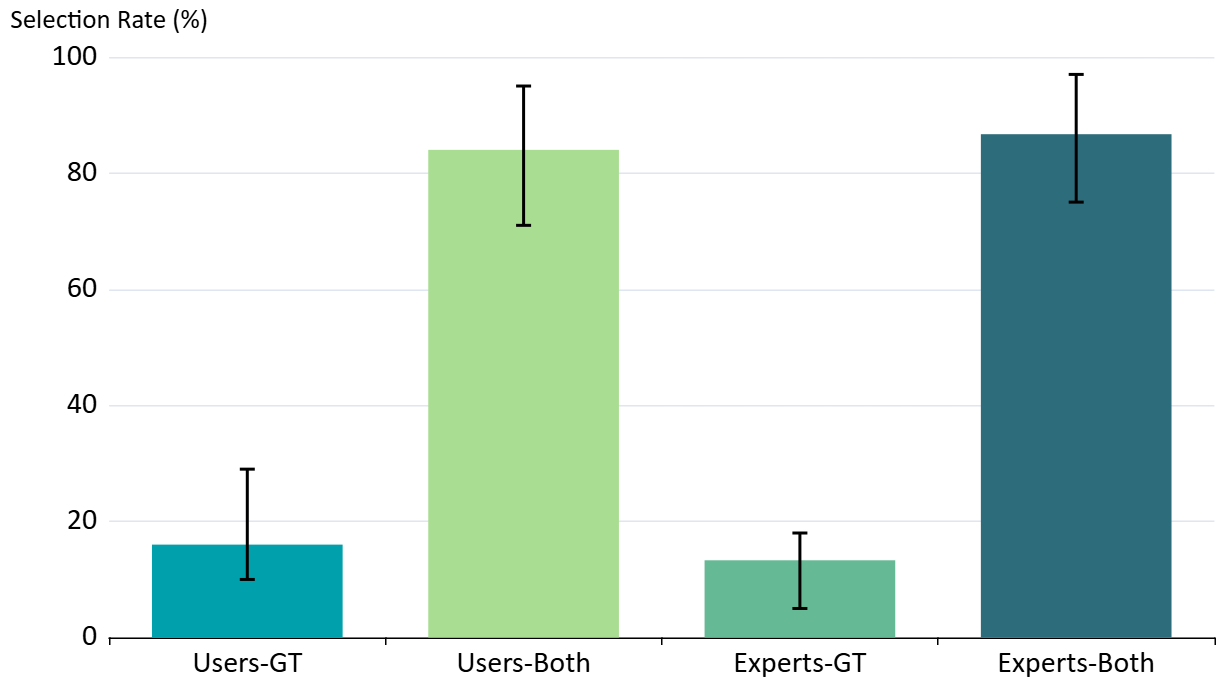}
	\caption{Results of user study.}
	\label{5.5}
\end{figure}

We invited ten participants with painting expertise and five art professionals to evaluate our results. Twenty painting sequences were randomly selected, each accompanied by the output generated by our model and the corresponding ground truth. The evaluators were asked to choose which result they preferred for each sequence, with the option to select both if they found them equally good. As shown in~\autoref{5.5}, both regular participants and domain experts consistently preferred the assessment results produced by our model or considered them comparable to the ground truth. This demonstrates the human-aligned nature of our method from both professional and non-professional perspectives.

\subsection{Visual Analysis}
\begin{figure}[]
	\centering
	\includegraphics[width=1\linewidth]{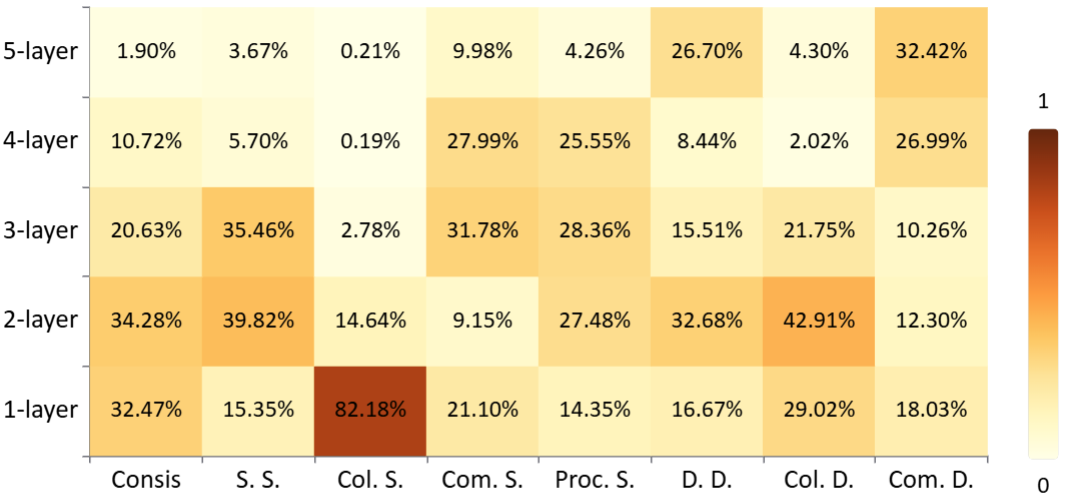}
	\caption{Heatmap of expert usage across different depths for each attribute.}
	\label{5.6}
\end{figure}

We visualize the internal routing behavior of our MoE model. We randomly select 1000 samples, and for each attribute, tokens with normalized contribution weights greater than 0.5 are identified as representative. We then analyze which experts these representative tokens are routed to in the final MoE layer. As shown in Figure~\ref{5.6}, the heatmap illustrates the expert usage frequency across different expert depths for each attribute. We observe that low-level attributes (e.g., color) tend to rely on shallower experts, while high-level attributes (e.g., composition) are routed to a broader range of expert depths. Overall, our MoE architecture exhibits balanced expert utilization and avoids the load imbalance issue, which contributes to improved model performance.

\section{Conclusion}
This paper addresses the challenging task of painting process assessment. To this end, we construct the PPAD dataset, which contains approximately 15,000 real paintings and 10,000 synthetic paintings, each annotated with eight attributes relevant to the painting process. Furthermore, we propose a novel model that incorporates time-aware angular correction in rotary positional encoding, along with hierarchical MoE modules enhanced by style awareness. Both quantitative experiments and user studies demonstrate that our method achieves state-of-the-art performance in painting process assessment. We also conduct ablation studies to evaluate the contribution of each component in our model.

In future work, we plan to further expand the scale and diversity of the PPAD dataset. Additionally, we aim to enrich the annotation of the painting process with linguistic descriptions, enabling a deeper understanding of the artist's skill level and providing more insightful feedback.

\begin{acks}
	The authors wish to acknowledge the support from NSSFC under Grant 22ZD05, NNSFC under Grant 62472178, and the Natural Science Foundation of Shanghai Municipality, China under Grant 24ZR1418300.
\end{acks}

\bibliographystyle{ACM-Reference-Format}
\bibliography{acmart}

\end{document}